%% file: shnn.tex
\documentclass[a4paper,1p]{elsarticle}

\usepackage{fancyhdr}

\usepackage{hyperref}
\usepackage{url}            

\usepackage[final]{changes}

\journal{Journal of Computational Physics}

\usepackage{numcompress}
\usepackage[utf8]{inputenc} 
\usepackage[T1]{fontenc}    

\usepackage{calc}
\usepackage{tikz}
\usetikzlibrary{shapes.geometric, arrows, arrows.meta, calc, intersections}

\usepackage{booktabs}       
\usepackage{amsfonts}       
\usepackage{nicefrac}       
\usepackage{microtype}      
\usepackage{xcolor}         

\usepackage{comment}
\usepackage{physics}
\usepackage[mathx]{mathabx}
\usepackage{amssymb,amsmath,amsthm}
\usepackage{slashed}
\usepackage{enumitem}
\usepackage{wrapfig}
\usepackage{float}
\usepackage{soul}
\usepackage{needspace}
\usepackage{subcaption}

\newcommand{\RR}[0]{\mathbb R}
\renewcommand{\phi}{\varphi}
\renewcommand{\epsilon}{\varepsilon}
\newcommand{\hatH}[0]{\smash{\widecheck H}}
\newcommand{\hatHse}[0]{\smash{\widecheck H}_\text{se}}
\newcommand{\hatHmp}[0]{\smash{\widecheck H}_\text{mp}}

\newtheorem{proposition}{Proposition}

\newtheorem*{corollary*}{Corollary}
\theoremstyle{remark}
\newtheorem*{remark}{Remark}
\newtheorem*{remarks}{Remarks}

\begin{document}

\begin{frontmatter}

\title{Symplectic Learning for Hamiltonian Neural Networks}

\author[sa,ens]{Marco David\corref{corauthor}}
\author[rennes]{Florian Méhats}

\address[sa]{SpaceAble (Paris, France)}
\address[ens]{École Normale Supérieure, PSL Research University, F-75005 Paris, France}
\address[rennes]{Univ Rennes, CNRS, IRMAR -- UMR 6625, F-35000 Rennes, France}
\cortext[corauthor]{\added{Corresponding author: \texttt{symplectic-hnn@ens.fr}}}

\begin{abstract}
	Machine learning methods are widely used in the natural sciences to model and predict physical systems from observation data. Yet, they are often used as poorly understood ``black boxes,'' disregarding existing mathematical structure and invariants of the problem. Recently, the proposal of Hamiltonian Neural Networks (HNNs) took a first step towards a unified ``gray box'' approach, using physical insight to improve performance for Hamiltonian systems.
	In this paper, we explore a significantly improved training method for HNNs, exploiting the symplectic structure of Hamiltonian systems with a different loss function. This frees the loss from an artificial lower bound. We mathematically guarantee the existence of an exact Hamiltonian function which the HNN can learn. This allows us to prove and numerically analyze the errors made by HNNs which, in turn, renders them fully explainable. Finally, we present a novel post-training correction to obtain the true Hamiltonian only from discretized observation data, up to an arbitrary order.
\end{abstract}


\end{frontmatter}

\thispagestyle{fancy}
\fancyhf{}
\renewcommand{\footruleskip}{4ex}
\fancyfoot[L]{\it\small Published by Elsevier in Journal of Computational Physics, \\ vol. 494 (2023), 112495, see \href{ 	
https://doi.org/10.1016/j.jcp.2023.112495}{doi:10.1016/j.jcp.2023.112495}.}
\fancyfoot[R]{\it\small \today}

\newpage
\section{Introduction}
\input{s1-intro}


\section{Theory}\label{sec:theory}
\input{s2-theory}


\section{Numerical Experiments}
\input{s3-experiments}


\section{Discussion and Conclusion}
\input{s4-discussion}


\section*{Acknowledgements and Disclosure of Funding}
The authors would like to express their deep gratitude to SpaceAble for sponsoring, supporting and encouraging this project. In particular, thank you to Issao Ueda, Arnaud Bellizzi, Louis Celier, Quentin Gueho and Julien Cantegreil for all their help. The authors further thank Philippe Chartier for interesting discussions and pointers in the right directions.

\bibliography{references}

\appendix

\section{Non-existence of a learnable Hamiltonian $\hatH$ for HNNs}\label{ap:non-existence-H}
\input{sa-proof}


\section{Model sizes and corresponding hyperparameters}\label{ap:training-parameters}

For each task and discretization time step $h$, we generated a dataset of $K$ points composed of two snapshots $(y_0, y_1)$ of the system, taken a time $\Delta t = h$ apart from each other. Further, we specified the size of a neural network with $L$ hidden layers of $M$ neurons, respectively, and then trained several models using the three different numerical integration schemes (forward Euler, symplectic Euler, implicit midpoint). The following table summarizes these choices. 

\begin{table}[!h]
	\centering
	\caption{Summary of model and dataset parameters for each task and time step $h$. Each cell specifies a network of $L$ hidden layers with $M$ neurons each, and a dataset of $K$ points, in the format $L, M; K$.}\label{tab:model-parameters}
	\vspace{.5\baselineskip}
	\begin{tabular}{l@{\hspace{2ex}}ccc}
		\toprule
		 & $h=0.8, 0.4, 0.2$ & $h=0.1$ & $h=0.05$ \\
		\midrule
		Spring ($\Omega_d = [-1, 1]^2$) & 1, 200; \phantom{10}2k & 2, 200; \phantom{10}4k & 3, 200; \phantom{0}10k \\
		Pendulum ($\Omega_d = [-\pi, \pi]^2$) & 1, 200; \phantom{10}2k & 2, 200; \phantom{10}4k & 3, 200; \phantom{0}10k \\
		Double Pend. ($\Omega_d = [-\pi, \pi]^4$) & 2, 400; 100k & 3, 600; 100k & 3, 600; 100k \\
		\bottomrule
	\end{tabular}
\end{table}


\newpage
\section{Analysis of the Hamiltonian error distribution in $\Omega_d$}\label{ap:hamiltonian-error-distribution}

\begin{figure}[H]
	\centering
	
	\parbox[m]{.3\linewidth}{%
	\subcaption{Harmonic Oscillator, forward Euler (HNN), $h=0.8,0.2,0.05$}
	}\hspace{2ex}\parbox{.65\linewidth}{%
	\includegraphics[width=\linewidth]{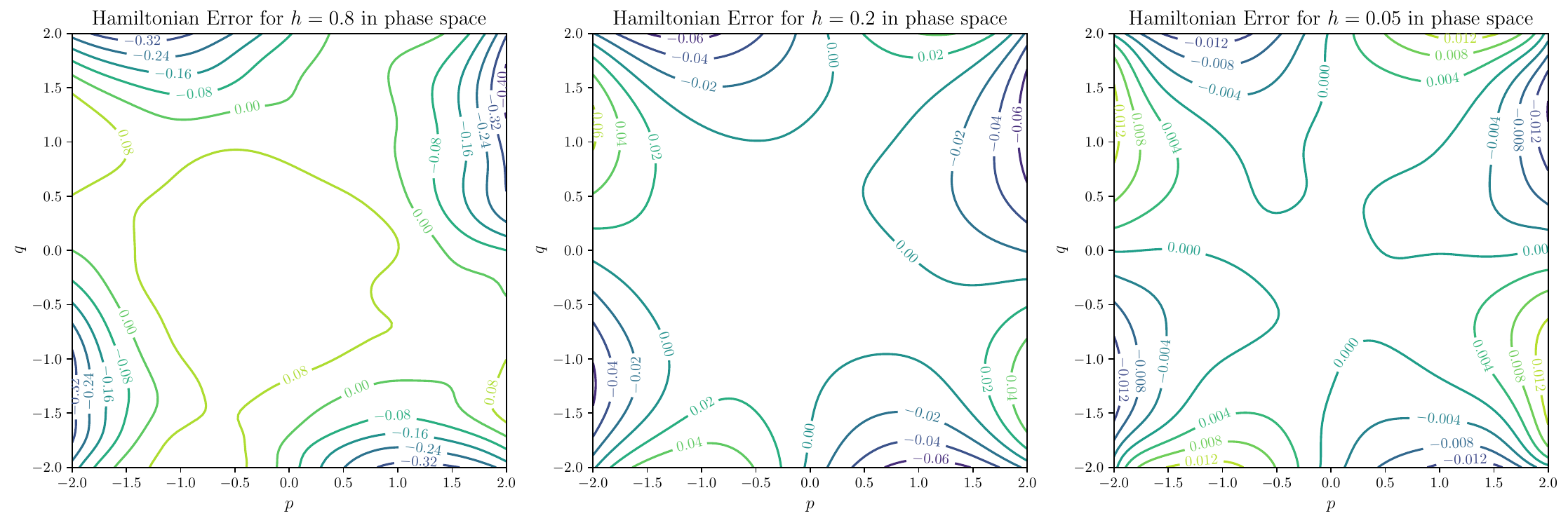}\\
	}
	
	\parbox[m]{.3\linewidth}{%
	\subcaption{Harmonic Oscillator, symplectic Euler, $h=0.8,0.2,0.05$}
	}\hspace{2ex}\parbox{.65\linewidth}{%
	\includegraphics[width=\linewidth]{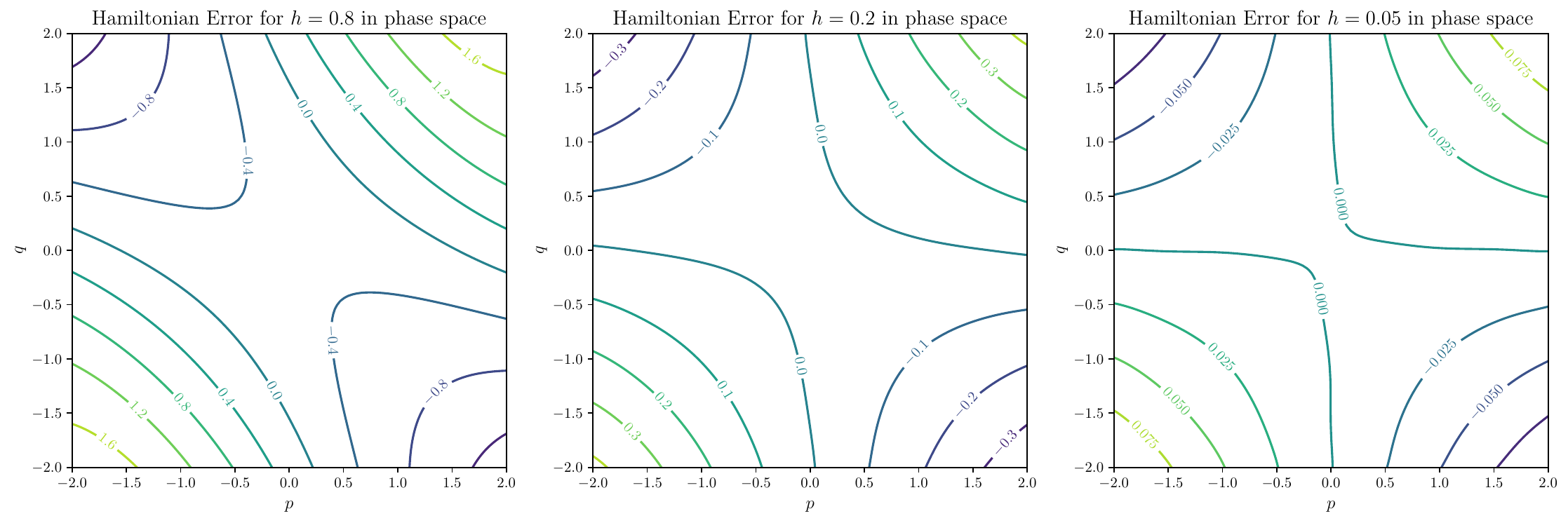}\\
	}
	
	\parbox[m]{.3\linewidth}{%
	\subcaption{Harmonic Oscillator, implicit midpoint, $h=0.8,0.2,0.05$}
	}\hspace{2ex}\parbox{.65\linewidth}{%
	\includegraphics[width=\linewidth]{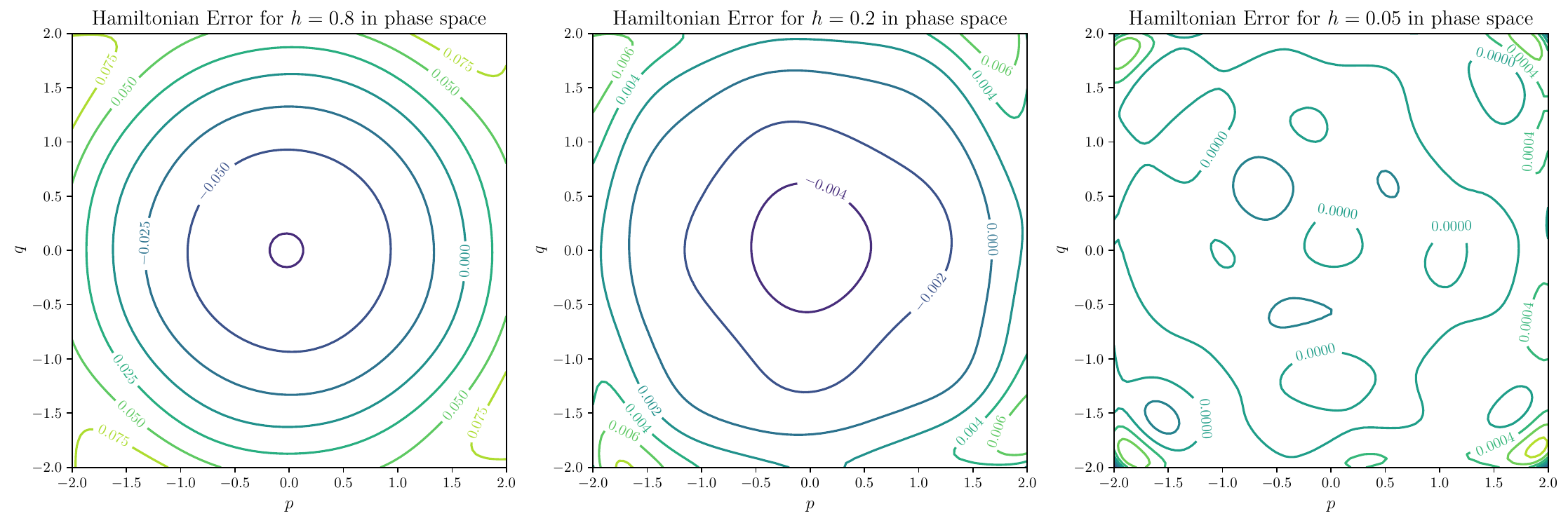}\\
	}

	\caption{Distribution of the error of the Hamiltonian $\hatH$ learned by an SHNN with respect to the true known Hamiltonian $H$. The different models were trained for the harmonic oscillator (Task~1) with the forward Euler, symplectic Euler and implicit midpoint methods. Shown are plots for the values $h\in \{0.8,0.2,0.05\}$.}\label{fig:c5}
\end{figure}

An analysis of the error distribution $\varepsilon_H(p, q)$ in phase space $\Omega_d$ proves very insightful (see Figure~\ref{fig:c5}). The below error distributions reveal a mixture of the shape of the true Hamiltonian as well as the geometric properties of the used numerical integration methods. Additionally, these distributions also confirm that the error is proportional to $h^p$. Most importantly, however, they show that the error becomes very large at the edge of $\Omega_d$ which motivates our choice of a restricted measuring region $\Omega_m \subsetneq \Omega_d$.

\clearpage
\section{Supplementary plots of trajectory predictions}\label{ap:dephasing}

\begin{figure}[h]
	\centering
	\begin{subfigure}{.49\textwidth}
		\includegraphics[width=\linewidth]{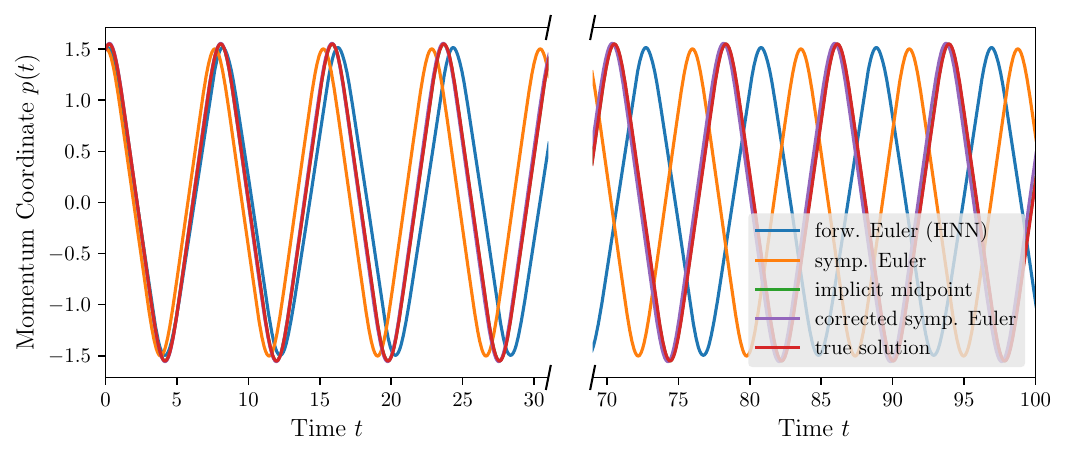}
		\caption{Time step $h=0.4$}
	\end{subfigure}
	\begin{subfigure}{.49\textwidth}
		\includegraphics[width=\linewidth]{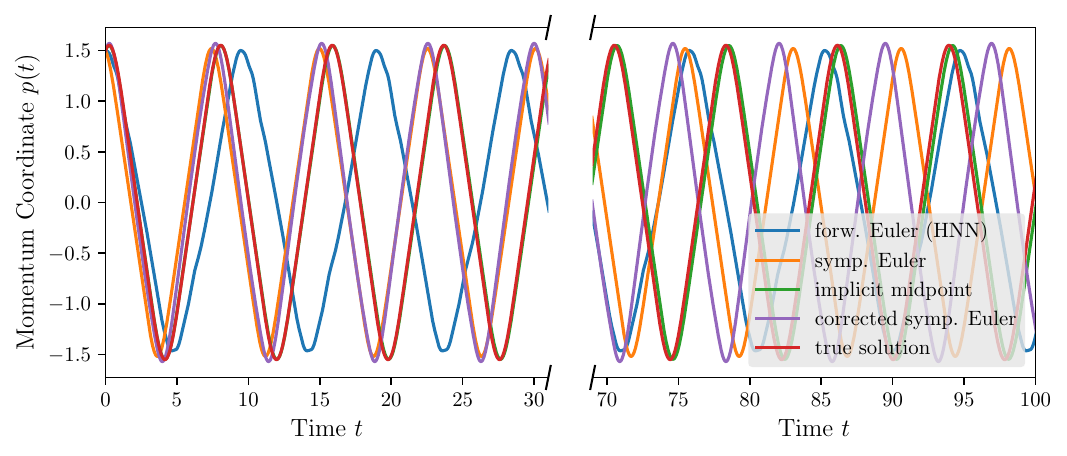}
		\caption{Time step $h=0.8$}
	\end{subfigure}
	\caption{Momentum coordinate of predicted trajectories of the pendulum system, with $h=0.4, 0.8$, starting from an initial position $p_0 = 1.5$ and $q_0 = -0.4$. In both plots, the prediction using the implicit midpoint method is almost completely hidden behind the true solution.}\label{fig:d6}
\end{figure}

For completeness, we also provide two exemplary plots of just one coordinate of trajectories predicted by (S)HNNs, in comparison with the true trajectory (see Figure~\ref{fig:d6}). This explains the origins of the mean squared $L^2$ error (MSE) as shown in Figure~\ref{fig:trajs} for the pendulum problem and (S)HNNs trained with $h=0.8$: There is a progressive dephasing between the model's prediction and the ground truth, which is why the MSE also oscillates on a large scale. In fact, the total phase difference is continuously growing with larger time (yet, it is only represented modulo $2\pi$ in the plots \textbf{C} of Figure~\ref{fig:trajs}).

\end{document}

%% file: s1-intro.tex

\begin{wrapfigure}{R}{0.5\textwidth}
	\centering
	\input{figs/scheme-hnn.tex}
	\caption{Exemplary architecture of an HNN with two hidden layers, due to \citet{greydanus}. The loss function, which optimizes the gradient $\grad \hatH(y)$ of the neural network with respect to its input $y$, is the principal object of study in this article.\label{fig:hnn-architecture}}
\end{wrapfigure}

Machine learning models provide a promising new method to learn and predict the behavior of physical systems. With much scientific attention devoted to complex systems such as the climate~\cite{PRL-analytic-constraints,de2018deepphysics}, fluid dynamics~\cite{ml-fluid-mechanics,wievel-neural-odes}, quantum mechanics~\cite{sellier-nn-for-quantum-physics}, or space weather~\cite{camporeale2019review,camporeale2018book}, machine learning might soon be a crucial tool for hybrid models~\cite{reichstein-deep-2019}, and maybe even replace some manually tuned models entirely~\cite{willard2020integrating}.

Many physical models can be described using systems of differential equations.
In particular, much general methodology to learn ordinary differential equations~(ODEs) has recently been developed \cite{brunton-sparse-ODEs,chen-neural-ode,nguyen2019emlike,regazzoni-ODEs}.
A specific and important class of physical models are Hamiltonian systems which are described by a single scalar function in phase space \mbox{$H : \Omega \subseteq \RR^{2n} \rightarrow \RR$}, often interpreted as the total energy of the system. The corresponding dynamics of the system for $y \in \Omega$ are described by Hamilton's equations~\cite{hamilton,ana-mech-textbook}
\begin{equation}\label{eq:hamilton}
	\dot y = J^{-1} \grad H(y), \qquad J = \mqty(0 & 1 \\ -1 & 0),
\end{equation}
that is a system of ODEs with the particular vector field $J^{-1}\grad H$. Here, $J$ is the so-called \emph{canonical symplectic matrix} and we also call the vector field the \emph{symplectic derivative} of $H$. Solutions of this system are physical trajectories $y(t)$ which necessarily have the property that $H(y(t))$ stays constant at all times~\cite{ana-mech-textbook}.

\citet{greydanus} have recently proposed a clever class of Hamiltonian Neural Networks~(HNNs) whose architecture (see Figure~\ref{fig:hnn-architecture}) engraves the mathematical properties of Hamilton's equations (notably their symplecticity). By asking the neural network to predict the Hamiltonian and then calculating its symplectic gradient using backpropagation, they were able to obtain trajectories in phase space that conserve the learned Hamiltonian over much longer periods of time than a baseline neural network which immediately learned the Hamiltonian vector field.

This work comprises three main contributions. In Section~\ref{subsec:analysis-hnn}, we conduct a mathematical analysis of HNNs which exhibits an intrinsic obstacle that prevents them from learning an exact Hamiltonian function. This hampers the best attainable accuracy of HNNs. Reinterpreting the loss function as the result of a numerical integration scheme, this obstacle realizes as an artificial lower bound on the loss. In a novel yet elementary proof (see \ref{ap:non-existence-H}), we show that this is due to the local error of the forward Euler integration method.

Section~\ref{subsec:shnn-introduction} then explores an improved training method for Hamiltonian Neural Networks which remedies this fact through a subtle change in the model's loss function. This approach has received several names but may simply be called \emph{Symplectic Hamiltonian Neural Networks}, or SHNNs. We prove that there always exists an exact function, the so-called modified Hamiltonian, that SHNNs can learn to arbitrary precision. In conjunction with \added{a thorough} experimental verification of this result (see Section~\ref{subsec:results}), we contribute to the explainability of general SHNNs.

Finally, Section~\ref{subsec:shnn-correction} introduces an entirely new post-training correction that allows to obtain the true Hamiltonian up to an arbitrary desired order. Using SHNNs and only discretized data of the true solutions, this approach exploits known properties of the modified Hamiltonian. Concretely, a Hamilton-Jacobi partial differential equation (PDE) is used to obtain a formal series expansion of the true Hamiltonian which can be used to correct for the introduced discretization error.

\subsection{Related Work}
Since the original proposal by \citet{greydanus} and the concurrent work by \citet{bertalan-chaos2019}, HNNs have generated much scientific interest. To name a few, generative~\cite{HGNs}, recurrent~\cite{Chen2020SymplecticRNN} and constrained~\cite{Zhong2020SymplecticODENet} versions, as well as Lagrangian Neural Networks~\cite{LNN} have been proposed.

\paragraph{Improvements of HNNs using symplectic integrators}
An adapted loss function for HNNs, derived from a symplectic numerical integration scheme, has been concurrently explored in several recent publications. \citet{zhu2020deep} use the implicit midpoint rule in an adapted loss and show that, in this case, the SHNN can learn an exact Hamiltonian given by the modified equation. Similarly, \citet{Chen2020SymplecticRNN} make a modification using the leapfrog integrator before proposing a recurrent, multi-step training method, but only for separable Hamiltonians. Neither paper uses the modified equation to correct the learned Hamiltonian. Independently, \citet{xiong2021nonseparable} use explicit higher-order symplectic methods during training, although they need to consider an augmented Hamiltonian on an augmented phase space with double the dimensionality.
Finally, \citet{dipietro-ssinn} use strong assumptions (including separability) to inform their architecture and train with a fourth-order symplectic integrator that is then explicit. However, they succeed in training with very small, sparse datasets.

\paragraph{Learning Hamiltonians from data using other architectures}
Beyond the immediate improvements of HNNs, there exist other proposals, notably other architectures, to learn Hamiltonians from data. Notably, \citet{Zhong2020SymplecticODENet} learn both a strongly parametrized and a general Hamiltonian in conjunction with the Neural ODE~\cite{chen-neural-ode} model. They also learn systems under the influence of external forces. Further, \citet{jin-sympnets} introduce a new architecture to learn any symplectic map and prove respective approximation theorems. They use SHNNs with the implicit midpoint rule as their baseline. \citet{taylornets} also directly predict a future state of the system, but they use a separable Hamiltonian in conjunction with a then explicit fourth-order symplectic method. Finally, \citet{chen-tao-icml} propose and analyze a method to exactly learn symplectic evolution maps which notably include general Hamiltonian dynamics.

%% file: figs/scheme-hnn.tex
\newdimen\XCoord
\newdimen\YCoord

\newcommand*{\ExtractCoordinate}[1]{\path (#1); \pgfgetlastxy{\XCoord}{\YCoord};}%
\newcommand*{\LabelCurrentCoordinate}[2]{\fill [#1] ($(\XCoord,\YCoord)$) circle (2pt) node [right] {#2}}%

\small
\begin{tikzpicture}[every text node part/.style={align=center}]
	\tikzstyle{arrow} = [->,>=stealth]
	
	\node[draw,circle,text centered,minimum height=1em] at (0, 4.2) (H) {$\phantom{i}\smash{\hatH}\phantom{i}$};
	
	\node[draw,rounded corners=1.8ex,text width=16em,text height=1em] at (0, 3) (L2) {};
	\ExtractCoordinate{$(L2.north)$}
	\coordinate (L2Y) at ($(0, \YCoord)$);
	
	\node[draw,rounded corners=1.8ex,text width=16em,text height=1em] at (0, 1.5) (L1) {};
	\ExtractCoordinate{$(L1.north)$}
	\coordinate (L1Y) at ($(0, \YCoord)$);
	
	\node[draw,rounded corners=.5ex,text centered,text width=5em,minimum height=2em] at (-1.2, 0) (IN) {$y = (p, q)$};
	\ExtractCoordinate{$(IN.north)$}
	\coordinate (INX) at ($(\XCoord, 0)$);
	
	\node[draw,dashed,rounded corners=.5ex,text centered,text width=5em,minimum height=2em] at (1.2, 0) (G) {$\grad {\widecheck H}(y)$};
	\ExtractCoordinate{$(G.north)$}
	\coordinate (GX) at ($(\XCoord, 0)$);

	
	
	\ExtractCoordinate{$(L1.south) - (IN.north)$}
	\draw [arrow] (IN.north) -- ($(IN.north) + (0, \YCoord)$) node [midway,left] {forw. pass};
	\ExtractCoordinate{$(L1.south) - (G.north)$}
	\draw [arrow,dashed] ($(G.north) + (0, \YCoord)$) -- (G.north);
		
	\ExtractCoordinate{$(L2.south) - (L1.north)$}
	\draw [arrow] ($(INX) + (L1Y)$) -- ($(INX) + (L1Y) + (0, \YCoord)$);
	\draw [arrow,dashed] ($(GX) + (L1Y) + (0, \YCoord)$) -- ($(GX) + (L1Y)$);
	
	\draw [arrow] ($(INX) + (L2Y)$) to[out=90,in=180] (H.west);
	\draw [arrow,dashed] (H.east) to[out=0,in=90] node [midway,right,xshift=1ex] {in-graph\\gradient} ($(GX) + (L2Y)$);
\end{tikzpicture}


%% file: s2-theory.tex

Consider a physical system whose state is described by a position vector $q \in \RR^n$ and a momentum vector $p \in \RR^n$. Abbreviate this state as $y = (p, q) \in \Omega \subseteq \RR^{2n}$. Written in position and momentum coordinates, Hamilton's equation~\eqref{eq:hamilton} reads
\begin{equation}\label{eq:hamilton-pq}
	\dot p = -\grad_q H(p, q), \qquad \dot q = \grad_p H(p, q).
\end{equation}
For example, in classical mechanics, the Hamiltonian can be written as the sum of kinetic and potential energy leading to a so-called separable Hamiltonian $H(p,q) = T(p) + V(q)$. Further, the kinetic energy of classical massive particles is $T(p) = p^2 / 2m$ such that Hamilton's equation reduces to Newton's second law for the acceleration $\ddot q$:
\begin{equation}
	\dot q = \frac{p}{m} \qq{and} \dot p = -\grad V(q) \qq{$\implies$} m\ddot q = -\grad V(q).
\end{equation}

If the Hamiltonian is known, the system \eqref{eq:hamilton} can be numerically integrated using a \emph{numerical integration method} $\Phi_h$ that maps $\hat y_i \mapsto \hat y_{i+1} = \Phi_h(\hat y_i)$, where $h$ is the integration time step. Let $\phi_t(y)$ denote the true flow of the system starting from the initial position $y$ at time $t=0$. The \emph{order} of a numerical method is then defined to be the largest exponent $\added{r}$ such that $\Phi_h(y) - \phi_h(y) = \mathcal O(h^{\added{r}+1})$.

For example, the simplest and most naive numerical method, the forward Euler method, simply updates $\hat y_i$ by $h$ times the ODE vector field evaluated at that point: $\Phi_h(\hat y_i) = \hat y_i + hJ^{-1}\grad H(\hat y_i)$. In position and momentum coordinates, this first-order method explicitly becomes
\begin{equation}
	\hat p_{i+1} = \hat p_i - h \grad_q H(\hat p_i, \hat q_i) \qq{and} \hat q_{i+1} = \hat q_i + h \grad_p H(\hat p_i, \hat q_i).
\end{equation}

A much more useful class of numerical methods are so-called \emph{symplectic methods}, defined by the fact that $\Phi_h(y)$ is a symplectic mapping (i.e. its Jacobian matrix $\Phi_h'(y)$ leaves the canonical symplectic form $J$ invariant). In two dimensions, these transformations preserve the area in phase space. Further, they preserve a scalar quantity close to the real Hamiltonian, called the \emph{modified Hamiltonian} (see ``Backward Error Analysis'' which may be called ``conjugate'' to the below analysis~\cite[Ch.~IX]{GNI}). Although symplectic methods also introduce an error of order $h^{\added{r}+1}$ at each step, due to their geometric nature, they are much better suited to integrate Hamilton's equation.

\subsection{Hamiltonian Neural Networks}\label{subsec:analysis-hnn}

In contrast to obtaining trajectories from a known Hamiltonian, the purpose of Hamiltonian Neural Networks (HNNs) \cite{greydanus} is to learn a Hamiltonian from data, composed of observed trajectories $y(t)$ which solve Hamilton's equation~\eqref{eq:hamilton}. More specifically, we consider a data point to be a couple $(y_0, y_1 = \phi_h(y_0))$ of two consecutive snapshots of the state of the system separated by a time $\Delta t = h$. Having two data points is crucial to have information about the evolution of the system and to calculate a finite difference approximation $(y_1-y_0)/h$ of the time derivative $\dot y$. Since any trajectory with $n$ data points can be split into $n-1$ such couples, we shall consider all our data to be in this form.

Denote the HNN by the function $\hatH(p, q)$ which implicitly also depends on all weights and biases of the chosen neural network, see Figure~\ref{fig:hnn-architecture} for a sketch of the architecture of HNNs. In the spirit of \citet{greydanus}, its loss function\footnote{Note that \citet{greydanus} used the analytic gradient of the true Hamiltonian as the target for most tasks, which yields a different mathematical problem, i.e. learning a known scalar function from its gradient. The present article only uses finite differences as would be obtained from real data.} for one data point $(y_0, y_1)$ is
\begin{equation}\label{eq:loss-hnn}
	\mathcal L_\text{HNN} = \norm\Big{\frac{y_1 - y_0}{h} - J^{-1}\grad \hatH(y_0)}^2_{L^2}
\end{equation}
which we shall rewrite into the form
\begin{equation}\label{eq:loss-hnn-rewritten}
	\mathcal L_\text{HNN} = h^{-2} \norm\Big{ y_1 - \smash{\underbrace{\qty(y_0 + hJ^{-1}\grad \hatH(y_0))}_{=\, \hat y_1}} }^2_{L^2}.
\end{equation}
\vspace{1ex} 

Now it becomes clear that, up to a constant factor of $h^2$, we are effectively integrating the HNN's prediction $\hatH$ using the forward Euler method and comparing the result $\hat y_1$ to the real observation $y_1$ in the loss function. We can do better than forward Euler!

\begin{remarks}\leavevmode
\begin{enumerate}[label={(\roman*)},nosep]
	\item In fact, using the forward Euler method here means that there does not even exist a function~$\hatH$ such that the loss could be identically zero everywhere in phase space\added{; a consequence of the converse Poincaré theorem \cite[Thm.~VI.2.6]{GNI}.} This impossibility realizes as a mismatch of order $h$ in the mixed second derivatives $\grad_{pq} \hatH$ and $\grad_{qp} \hatH$ of the function to be learned, \added{as explained in \ref{ap:non-existence-H}.} \added{An artificial lower bound for the loss is thus introduced, frustrating the training procedure.}
	\item One may argue that, instead, the true derivative $\dot y$ should be more accurately approximated by higher-order difference quotients. Yet, this would only involve operations on the dataset without the HNN itself, so this point of view is rather limited. One would not be able to exploit the real system's symplecticity as explained next.
\end{enumerate}
\end{remarks}

\subsection{Symplectic methods to the rescue}\label{subsec:shnn-introduction}
According to the theory of geometric numerical integration~\cite{GNI}, the forward Euler method should be replaced by a symplectic method. The two simplest symplectic methods are the \emph{symplectic Euler method}
\begin{equation}\label{eq:symp-Euler}
	p_1 = p_0 - h \grad_q H(p_1, q_0), \qquad q_1 = q_0 + h \grad_p H(p_1, q_0)
\end{equation}
and the \emph{implicit midpoint rule}
\begin{equation}\label{eq:midpoint}
	y_1 = y_0 + h J^{-1} \grad H\qty\Big(\frac{y_0 + y_1}{2}).
\end{equation}
Both methods are implicit and are obtained by only changing the point of evaluation of the Hamiltonian vector field. Abstracting the choice of a specific integration scheme by a function $s(y_0, y_1)$ (this covers all methods of interest in this article), we obtain the loss function for Symplectic Hamiltonian Neural Networks (SHNNs):
\begin{equation}\label{eq:loss-shnn}
\begin{gathered}
	\mathcal L_\text{SHNN} = \norm\Big{\frac{y_1 - y_0}{h} - J^{-1}\grad \hatH(s(y_0, y_1))}^2_{L^2} \\
	\text{(provided that $s(y_0, y_1)$ gives rise to a symplectic integration method)}
\end{gathered}	
\end{equation}

Using a symplectic method ensures that there exists a function $\hatH$, which the SHNN can theoretically learn to an arbitrary precision (assuming that the neural network is large enough and that the dataset sufficiently covers the input space), as encapsulated by the following proposition. \added{We shall henceforth refer to $\hatH$ as the \emph{(forward) modified Hamiltonian}.} Note that the solutions of a general Hamiltonian system are not necessarily well-behaved for long times, so all following theoretical results are \added{non-trivial and meaningful only for not too large} $h$.

\begin{proposition}\label{prop:1}
	Let $H : \added{\Omega \subseteq} \RR^{2n} \rightarrow \RR$ be a smooth Hamiltonian and $\dot y = J^{-1} \grad H(y)$ be the corresponding Hamiltonian system. Fix $h>0$. Denote the true flow of the system after time $h$ by $\phi_h$, \added{and write} $\phi_h(y_0) =: y_1 = (p_1, q_1)$ \added{wherever this is well-defined for some initial condition $y_0 = (p_0, q_0)$}. Then,
	\begin{enumerate}[label={(\alph*)}]
		\item (symplectic Euler) there exists a \added{locally-defined} smooth function $\hatHse$ such that
		\begin{equation}
			J^{-1}\grad \hatHse (p_1, q_0) = \frac{y_1 - y_0}{h}
		\end{equation}
		for all $y_0 \in \added{\Omega}$ where $y_1 = \phi_h(y_0)$ is well-defined.
		
		\item (implicit midpoint) there exists a \added{locally-defined} smooth function $\hatHmp$ such that
		\begin{equation}
			J^{-1}\grad \hatHmp \qty\Big(\frac{y_1 + y_0}{2}) = \frac{y_1 - y_0}{h}
		\end{equation}
		for all $y_0 \in \added{\Omega}$ where $y_1 = \phi_h(y_0)$ is well-defined.
	\end{enumerate}
\end{proposition}

\begin{proof}
	This is exactly the result of Lemma 5.3 in \citet[Ch.~VI]{GNI}, using $\hatHse = \frac{1}{h} S^1$ or $\hatHmp = \frac{1}{h} S^3$ \added{and noting that $\phi_h$ is a smooth symplectic transformation close to the identity.}
\end{proof}

\begin{remarks}\leavevmode
\begin{enumerate}[label={(\roman*)},nosep]
	\item A more general result of existence for any general symplectic integration method is given by \citet[Sec.~2.2]{chartier} using so-called B-series, a generalized, formal Taylor expansion. What they call the ``modified differential equation,'' written with a modified Hamiltonian, is exactly what an SHNN learns.
	\item Note that when numerically solving ODEs, implicit methods like the symplectic Euler method and midpoint rule normally require fixed point iterations at each step. However, during the training of HNNs we are solving the reverse problem for which the true trajectories $(y_0, y_1)$ are already known. This means that symplectic training with implicit methods has the same computational cost as training with explicit methods.
	\item One property of the learned Hamiltonian should be noted: \emph{Integrating the \added{respective} Hamilton equation of the \added{modified} $\hatH$ with that same method and same time step $h$ as used during training, one will obtain the true flow $\phi_H^h$ of the real Hamiltonian.} This is the statement of Theorems~5.7 and 5.8 of \cite[Ch.~VI]{GNI} (noting that the necessary assumptions are satisfied due to Proposition~\ref{prop:2} below). However, there is no free lunch: The true intermediate states of the system will not be accessible nor predictable with this method. In fact, reducing the integration time step in this case will worsen the quality of the trajectory.
\end{enumerate}
\end{remarks}

\subsection{Correction of the learned Hamiltonian}\label{subsec:shnn-correction}
What's more, we can derive a formal series for the modified Hamiltonian $\hatH$ learned by an SHNN in terms of the real Hamiltonian $H$ and its derivatives, depending on the used integration method. This allows us to not only understand exactly what our model learns and where it draws its predictive power from, but also to correct the model after training to an arbitrary oder. This way, we can learn the real, physical Hamiltonian $H(p, q)$ to arbitrary precision, purely from discretized snapshots of trajectory data, without any information about the true gradients or vector fields.

Mathematically, Hamilton's equation is the characteristic equation of the Hamilton-Jacobi partial differential equation (PDE), obtained by considering the flow of the true Hamiltonian system after a variable time $t \in \RR$. Rendering explicit the fact that the modified Hamiltonian naturally depends on the time step $h$ fixed in Proposition~\ref{prop:1}, we write $\hatH(p, q, t=h)$. Since the flow is a smooth function of time when it exists, $\hatH$ will also be smooth in all its variables. Calculating its time derivative, this leads to the following.

\begin{proposition}\label{prop:2}
	Let $H : \added{\Omega \subseteq} \RR^{2n} \rightarrow \RR$ be a smooth Hamiltonian. For both cases of Proposition~\ref{prop:1}, there exists a neighborhood of $t=0$ where the respective time-dependent modified Hamiltonian $\hatH(p, q, t)$ solves a Hamilton-Jacobi PDE. Explicitly,
	\begin{enumerate}[label={(\alph*)}]
		\item (symplectic Euler) there exists an open neighborhood $U \ni 0$ such that $\forall t \in U$, 
		\begin{equation}
			\pdv{t}\qty\big(t\hatHse(p, q, t)) = H\qty(p, q + t \pdv{\hatHse}{p} \, \added{(p, q, t)})
		\end{equation}
		\item (implicit midpoint) there exists an open neighborhood $U \ni 0$ such that $\forall t \in U$, 
		\begin{equation}
			\pdv{t}\qty\big(t \hatHmp(y, t)) = H\qty(y + \frac{t}{2} J^{-1} \grad_y \hatHmp (y, t))
		\end{equation}
	\end{enumerate}
	Integrating either of these relations from $0$ to $h \in U$ and Taylor expanding the right-hand side generates a formal power series in $h$ whose coefficients depend on $H$ and its (partial) derivatives.
\end{proposition}

\begin{proof}
	This is a direct calculation using generating functions~\cite[Sec.~VI.5.3]{GNI}.
\end{proof}

This proposition allows the computation of $\hatH$ learned by an SHNN if the Hamiltonian $H$ of the original problem is known (see also~\cite[Sec.~VI.5.4]{GNI}). However, in practice, we will want to reconstruct $H$ from the modified $\hatH$ learned from data. Hence, the formal power series needs to be inverted to the desired order. Since the series is purely formal, this can be easily done with the help of symbolic computation programs. For the symplectic Euler method, abbreviating $H = H(p,q)$ and $\hatH = \hatHse(p, q, h)$, this yields
\begin{equation}\label{eq:correction-symp-euler}
\begin{split}
	H = \hatH &- \frac{h}{2} \grad_p \hatH \cdot \grad_q \hatH \\
	&+ \frac{h^2}{12}\qty\Big(\grad_{pp} \hatH (\grad_q \hatH)^2 + 4\grad_{pq} \hatH(\grad_p \hatH,\grad_q \hatH) + \grad_{qq} \hatH (\grad_p \hatH)^2) + \mathcal O(h^3).
\end{split}
\end{equation}
Similarly, for the implicit midpoint method, one obtains
\begin{equation}\label{eq:correction-midpoint}
	H = \hatH - \frac{h^2}{24} \grad^2 \hatH \qty(J^{-1} \grad \added{\hatH}, J^{-1} \grad \added{\hatH}) + \mathcal O(h^4).
\end{equation}

\added{Both of these equations are precisely the standard series which express the \emph{(backward) modified Hamiltonian} of a Hamiltonian system via Backward Error Analysis. This calculation renders explicit how our analysis of SHNNs is conjugate to this well-known method: The backward modified Hamiltonian of the learnt (forward modified) Hamiltonian $\hatH$ is again the true Hamiltonian $H$. We shall refer to terms from these series as \emph{corrections} to the learnt Hamiltonian, reducing the error caused by our discrete data.}

\medskip

\added{Aside from the question which function can be learned in principle, one may ask how well a given network architecture approximates the target $\hatH$. As one example of recent theoretical work on this question (see also~\cite{jin-sympnets,lu-approximation-relu}), \citet{deryck} have studied fully connected neural networks with two hidden layers and a $\tanh$ activation function.}\footnote{\added{This makes the neural network smooth and also exactly matches the architecture used for most of our numerical experiments.}} \added{The following corollary uses their results on the universal approximation of Sobolev functions by such an architecture.}

\newcommand{\Linfty}[0]{{L^\infty(\Omega_m)}}
\begin{corollary*}
	\added{Let $H : \Omega \subseteq \RR^{2n} \rightarrow \RR$ be a smooth Hamiltonian system. Then, an SHNN $\mathfrak H$ with two hidden layers of $M$ neurons each and a $\tanh$ activation function, trained with a symplectic method of order $r$ and hyperparameters $\vec \beta$ on $K$ data points separated by a time step $h$, makes the following error on a compact smooth region $\Omega_m \subset \Omega$. For every $s \in \mathbb N$, it holds that}
	\begin{equation}
		\norm{\mathfrak H - H}_\Linfty \leq C_1(r, \hatH) h^r + C_2(n, s, \hatH)M^{-s} + \delta(K, M, \vec \beta, \ldots).
	\end{equation}
\end{corollary*}

\begin{remark}
	\added{The constants $C_1$ and $C_2$ depend, respectively, on the $W^{r,\infty}(\Omega_m)$ and $W^{s,\infty}(\Omega_m)$ Sobolev norms of the modified Hamiltonian.}\footnote{\added{Moreover, note that the network is only trained on derivatives of its output, hence it is able to learn $\hatH$ solely up to some global bias, as corrected for later in eq.~\eqref{eq:h-err}. We assume this global bias to be zero for the corollary above to make sense.}}
\end{remark}

\added{The corollary is proven with the triangle inequality. The first term captures the error between the modified and true Hamiltonians $\hatH$ and $H$ as developed earlier in this section. The second term captures the error in approximating Sobolev functions by a fully-connected tanh neural network which consists of iterated instances of $\tanh(Ax + b)$, following the theory developed by \citet{deryck}. The third term captures the learning error between the best possible approximation of $\hatH$ and the real network $\mathfrak H$, due to the data set, the chosen optimizer and its hyperparameters. The following section operationalizes this result.} 

%% file: s3-experiments.tex

Using a numerical implementation\footnote{Our source code, synthetic datasets and pre-trained models are openly available at: \url{https://github.com/SpaceAbleOrg/symplectic-hnn}.} of SHNNs, we have tested all of the above theory on three tasks of varying difficulty. The first two tasks were also treated by \citet{greydanus} in the original proposal of HNNs.

\paragraph{Task 1: The Harmonic Oscillator} The ``Hello World'' of Hamiltonian systems is the harmonic oscillator with non-dimensionalized Hamiltonian $H(p, q) = \frac{1}{2}p^2 + \frac{1}{2}q^2$. This simple system is of fundamental importance in physics as it is described by a quadratic potential (linear force). It models e.g. a mass on a spring, where $q$ is the displacement from equilibrium and $p$ the momentum.

\paragraph{Task 2: The Non-linear Pendulum} Another important system is the non-linear ideal pendulum with non-dimensionalized Hamiltonian $H(p, q) = \frac{1}{2}p^2 + (1 - \cos q)$. Instead of a linear restoring force, the gravitational force is now proportional to $\sin q$, where $q$ models the angle of the pendulum measured from equilibrium and $p$ consequently models the angular momentum.

\paragraph{Task 3: The Double Pendulum} As a non-separable system of dimension 4 with chaotic dynamics, the double pendulum presents the most challenging task. Its non-dimensionalized Hamiltonian
\begin{equation}
	H(p_1, p_2, q_1, q_2) = \frac{\frac{1}{2}p_1^2 + p_2^2 - p_1 p_2 \cos (q_1 - q_2)}{1 + \sin^2(q_1 - q_2)} - 2 \cos q_1 - \cos q_2
\end{equation}
describes two equal masses at angles $q_{1,2}$ and angular momenta $p_{1,2}$. The first mass is attached to a fixed point, and the second mass is attached to the first, using stiff massless rods of equal length.

\subsection{Methods}\label{subsec:methods}

For each task, four different models were trained based on the loss function~\eqref{eq:loss-shnn}, for different choices of the scheme function $s = s(y_0, y_1)$ and post-training correction: forward Euler, symplectic Euler, implicit midpoint, and corrected symplectic Euler.

The forward Euler scheme $s = y_0$ replicates HNNs \cite{greydanus} trained with discretized data and represents our baseline. The symplectic Euler scheme $s = (p_1, q_0)$ is a symplectic method of order 1 whereas the implicit midpoint rule \mbox{$s = (y_0 + y_1)/2$} is a symplectic method of order 2. Finally, we also trained an SHNN with the symplectic Euler method but afterwards corrected its Hamiltonian using $\added{\mathfrak H} - \smash{\frac{h}{2}} \grad_p \added{\mathfrak H} \cdot \grad_q \added{\mathfrak H}$, obtaining a Hamiltonian correct up to second order.

For each task, we defined a bounded subregion $\Omega_d$ of the full phase space $\Omega$ to generate the data from. Given a fixed time step $h > 0$, we generated a dataset of $K$ data points. Each point is given by a couple $(y_0, y_1)$ where $y_0$ is a random initial state chosen uniformly from $\Omega_d$ and $y_1 = \phi_h(y_0)$ represents a snapshot of the system's true solution at a time $h$ later. Note that friction was neglected for all tasks; in fact, the architecture of HNNs prevents them from learning any change of the total energy with time. The full data set was separated using a test split of 20\%.

We used fully connected neural networks with a $\tanh$ activation function, $L$ hidden layers and $M$ neurons per hidden layer for all tasks. All models were trained with the AdamW optimizer \cite{Adam,AdamW} as implemented in PyTorch~\cite{pytorch} using default coefficients, a learning rate of $10^{-3}$ and weight decay of $10^{-2}$, for 5000 epochs without mini-batches (i.e. batch size = $K$). Only the model with the best test loss was saved after training. Table~\ref{tab:model-parameters} in \ref{ap:training-parameters} summarizes the different choices of model and dataset size for all tasks.  Training was performed on a single GPU (Nvidia Tesla K80, cloud hosted) using the CUDA framework version 11.2~\cite{cuda-toolkit} as integrated in PyTorch~\cite{pytorch}.

The test and train $L^2$ losses were tracked per epoch while training the models. Afterwards, we measured our principal metric: the average error $\epsilon_H$ of the learned Hamiltonian over a region $\Omega_m \subset \Omega_d$ of phase space. This quantity was measured as
\begin{equation}\label{eq:h-err}
	\smash{\epsilon_H = \Big\langle \abs\big{\added{\mathfrak H} - H - \ev{\added{\mathfrak H} - H}_{\Omega_m}} \Big\rangle_{\Omega_m}},
\end{equation}
where the mean difference between $\mathfrak H$ and $H$ is removed inside the absolute value because the neural network only learns \added{(an approximation of)} $\hatH$ up to some \added{global} constant. \added{Intuitively, one would to like to remove this constant} by evaluation at a single point as $\added{\mathfrak H} - H - (\added{\mathfrak H}(\added{y_*}) - H(\added{y_*}))$. However, this would add the local error at $y=\added{y_*}$ to the function everywhere, whereas the mean does not suffer from any locality issues. \added{In particular, this measure $\varepsilon_H \leq \norm{\mathfrak H - H}_\Linfty$ as estimated in our Corollary in Section~\ref{sec:theory}.}

As an additional metric, we roll out long-term predictions of the trained models from random initial points, using the explicit Runge-Kutta method Dormand and Prince of order 5(4)~\cite{rk45} implemented in the \texttt{scipy.integrate} module~\cite{scipy}. Those trajectories are analyzed in two different fashions. Analyzing the shape of their trajectories in phase space, following the level curves of the modified Hamiltonian, provides insight into this modification with respect to the true Hamiltonian. Independently, measuring the mean squared $L^2$ error (MSE) between the long-term predictions of our models and the true solution provides insight into the quality of the predictions.

\medskip
\begin{remark}
The measuring region $\Omega_m$ was chosen as the hypercube centered and contained within the data region $\Omega_d \subseteq \RR^{2n}$, with side lengths divided by $\smash{\sqrt{2}}$. The average error was not measured directly on $\Omega_d$ because our models perform drastically worse close to the boundary of $\Omega_d$ (see \ref{ap:hamiltonian-error-distribution}).
\end{remark}

\subsection{Results}\label{subsec:results}

\begin{figure}
	\centering
	\includegraphics[width=\linewidth]{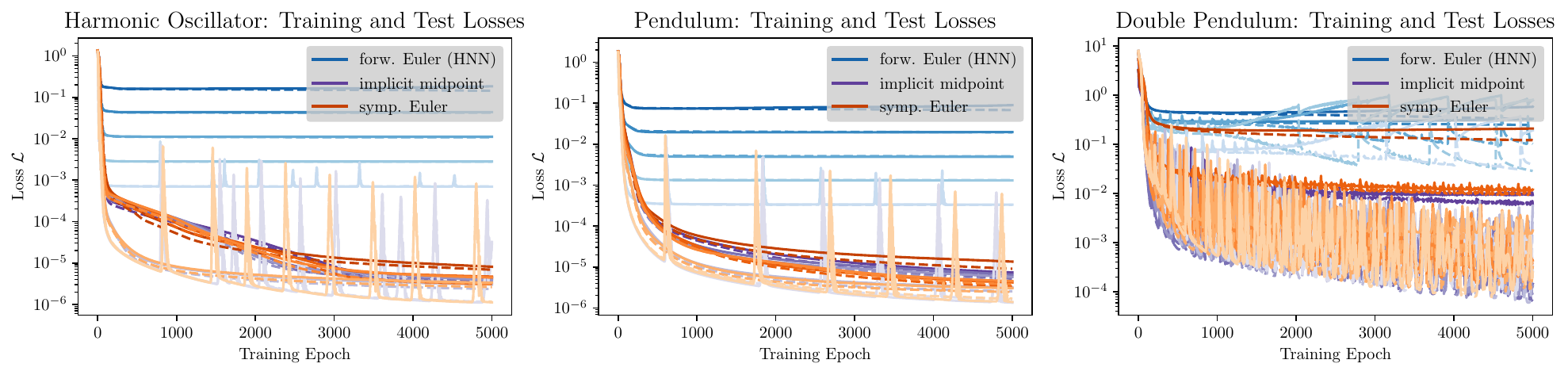}
	\caption{Training (dashed lines) and testing (solid lines) losses as a function of the training epoch for the three chosen tasks, the different integration methods and different discretization time steps $h \in \{0.05, 0.1, 0.2, 0.4, 0.8\}$. For each method, the darkest shade of its color corresponds to the largest $h=0.8$ and the lightest shade to the smallest $h=0.05$.}\label{fig:losses}	
\end{figure}

\begin{figure}[t]
	\centering
	\includegraphics[width=\linewidth]{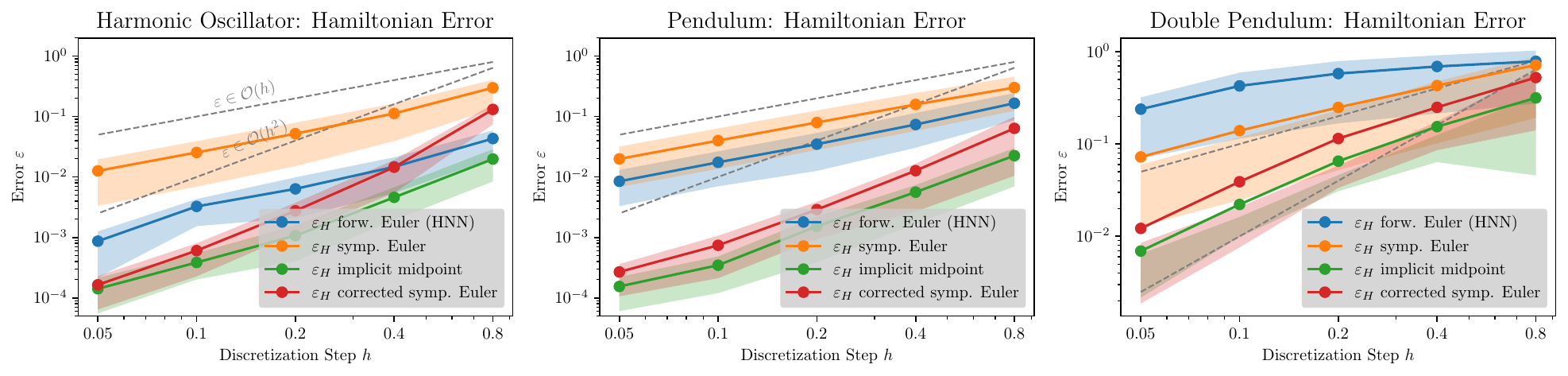}
	\caption{Average error of the learned Hamiltonian $\epsilon_H$ as a function of the discretization time step $h$, for the three chosen tasks and the different integration methods ($N=2000$). For each point the mean (solid marker) and quartiles (transparent region) are plotted, i.e. 50\% of all data points respectively lie inside the colored transparent regions. Note that, for every point, the standard error of the mean is too small to be visible. Two reference lines $\epsilon = h$ and $\epsilon = h^2$ have been added in grey.}\label{fig:h-err}
\end{figure}

The SHNN models trained well on all datasets, with only minimal overfitting. Figure~\ref{fig:losses} shows the training and test loss as a function of the training epoch, for an HNN trained with the forward Euler scheme and two SHNNs trained with the symplectic Euler and implicit midpoint schemes, respectively. Remarkable is the fact that the HNN losses plateau very quickly and, depending on the chosen time step $h$, do not descend below a certain threshold. These lower bounds of the squared $L^2$ loss are proportional to $h^2$ which confirms the theoretical result of \ref{ap:non-existence-H}. The fact that they vary by a constant factor realizes as a constant difference on the logarithmic scale. Contrarily, for the trained SHNNs and all three tasks, the losses descend independently of $h$ down to a level of numerical accuracy.

Further, we analyzed the average error of the learned Hamiltonian, which allows to draw inferences on the order of the used numerical method. The averages of equation~\eqref{eq:h-err} were calculated using $N=2000$ points uniformly drawn from $\Omega_m$. The mean and quartiles are shown in Figure~\ref{fig:h-err} on a double logarithmic scale, which means that an error of order $h^p$ realizes as a straight line with slope $p$. This figure shows that using the forward or symplectic Euler methods yields an error of order $h$, and that using the implicit midpoint method yields an error of order $h^2$ as expected. Further, they confirm that the post-training correction to the SHNN trained with the symplectic Euler method also yields an error of order $h^2$.

\begin{figure}
	\centering
	\includegraphics[width=\linewidth]{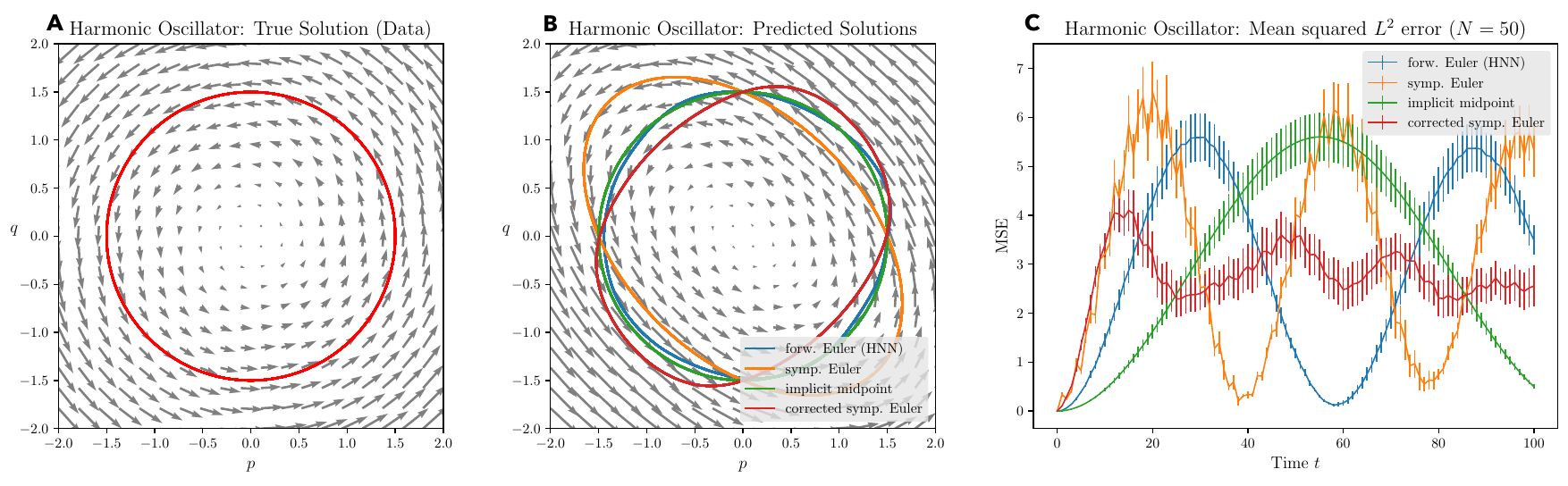}
	\includegraphics[width=\linewidth]{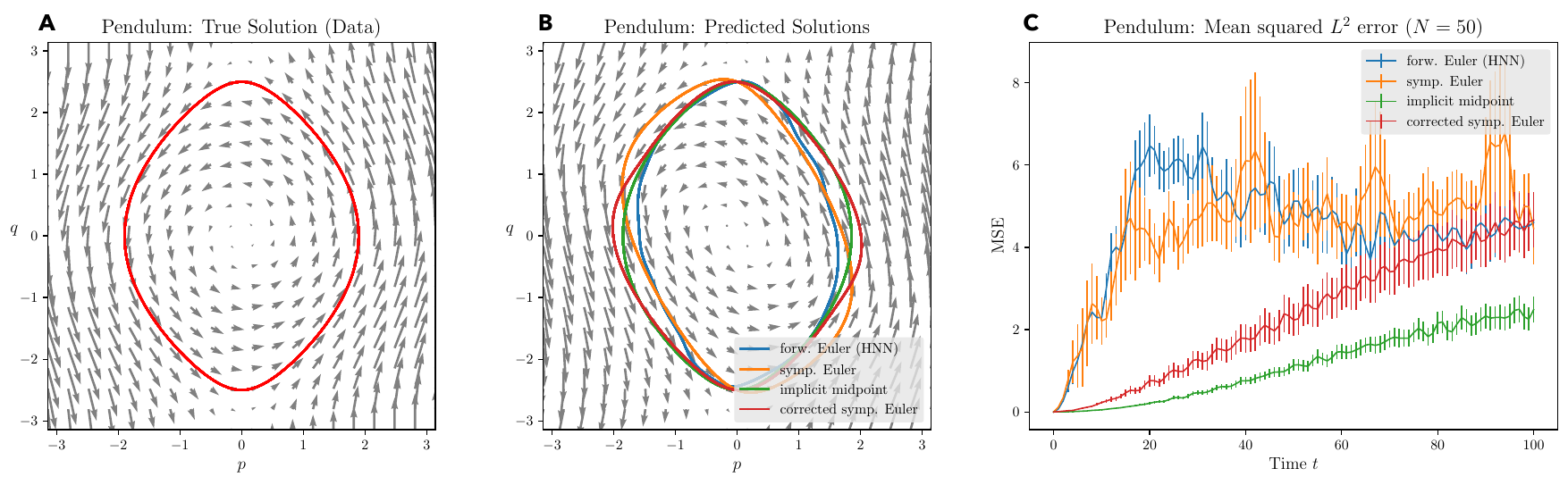}
	\caption{Analysis of individual trajectories of the pendulum and harmonic oscillator, with different SHNNs, trained with a discretization step $h=0.8$. The explicit trajectories drawn in phase space started at $y_0 = (0, 1.5)$ for the harmonic oscillator and $y_0 = (0, 2.5 \text{ rad})$ for the pendulum. The displayed regions of phase space are $\Omega_d$ in each case. \textbf{A}: True vector field and solution. \textbf{B}: Vector field as learned by an SHNN (trained with symplectic Euler) and predicted trajectories with all methods. \textbf{C}: Long-time results. MSE and standard error of the mean of $N=50$ trajectories, with initial points drawn randomly from $\Omega_m$, under the additional restriction that the system does not have enough energy to leave $\Omega_d$; since our models were not trained outside of $\Omega_d$.}\label{fig:trajs}
\end{figure}

Finally, Figure~\ref{fig:trajs} shows exemplary long-term trajectories for the spring (Task 1) and non-linear pendulum (Task 2) and a large step $h=0.8$, which makes the differences between the used integration schemes well visible. Several observations can be made. First, while the true Hamiltonian vector field is in general well learned by the SHNN, relatively large errors are visible at the edge of $\Omega_d$. Second, training with the implicit midpoint rule later predicts the most accurate trajectories, followed by the corrected symplectic Euler method. This is the case both in the shape of the trajectory as well as in the long-term MSE. Third, SHNNs with the symplectic Euler method learn a highly eccentric shape in phase space. This reflects the asymmetry of the method due to the evaluations at $s = (p_1, q_0)$, and in the case of the harmonic oscillator, this explains the perfect ellipse that results. Contrarily, the symmetric implicit midpoint method does not show this behavior. Fourth, correcting after training with symplectic Euler does improve the result (especially in the MSE) but ``overshoots'' the goal in phase space --- the corrected ellipse is eccentric in the opposite direction, as expected from the alternating signs in equation~\eqref{eq:correction-symp-euler}.

It is to be noted that the baseline HNN performs exceptionally well on Task~1 due to its simplicity. However, as Figure~\ref{fig:h-err} shows, too, the more complex the model, the worse the performance of an HNN trained with the forward Euler method. We shall also point to \ref{ap:dephasing} which allows to better understand the oscillations of the MSE.

%% file: s4-discussion.tex

The experimental results presented above confirm the theory developed in Section~\ref{sec:theory}. The loss curves of Figure~\ref{fig:losses} experimentally show that there is no exact function which can be learned by an HNN due to a mismatch depending on the chosen discretization step $h$. The errors of the learned Hamiltonians plotted as a function of $h$ in Figure~\ref{fig:h-err} demonstrate first that HNNs inherit their order from the numerical method used during training. Second, they show that post-training corrections indeed allow one to obtain higher orders by exploiting the formal series expansion obtained from the Hamilton-Jacobi PDEs in Proposition~\ref{prop:2}.

This last point deserves further discussion. Up to our knowledge, for general (i.e. non-separable) Hamiltonian systems, there do not exist other methods that allow a non-recurrent SHNN to efficiently learn the modified Hamiltonian accurately to any order larger than 2. If the dataset is of the form $\{(y_0, y_1)_i\}$, symplectic methods of higher orders are generally implicit \emph{multi-stage} methods\footnote{Explicit symplectic methods for non-separable Hamiltonians do exist, too, although they require an augmented phase space~\cite{tao-symplectic} twice the size of the physical phase space. For these methods, it is a priori not clear that an exact modified Hamiltonian exists in physical phase space.}, i.e. they use intermediate points like $y_{1/2}$ which are only implicitly determined themselves (unless the Hamiltonian is separable). Training with such a method means solving an implicit equation determined by the model for each data point and afterwards calculating the gradient of this operation --- a huge computational cost! In contrast, our post-training method only makes predictions slightly more expensive. Per time step, the computational cost is larger by a constant factor due to the extra evaluations of the model's derivatives, according to the correction formulas~\eqref{eq:correction-symp-euler} and~\eqref{eq:correction-midpoint}. Better yet, one may even hard-code any given correction to a model once trained.

Finally, we wish to note some observations for the double pendulum system. In Figures~\ref{fig:losses} and~\ref{fig:h-err}, the loss curves and average errors for $h=0.4$ and $h=0.8$ do not follow the theoretical expectation. We suspect that this is because the coarse grained data for these step sizes does not fully capture the (chaotic) dynamics of the system, and hence prevents accurate training. Additionally, for all values of $h$, the fact that the mean error in Figure~\ref{fig:h-err} lies outside the middle two quartiles shows that the error distribution is highly skewed, even after restriction to $\Omega_m$.

\subsection{Limitations}\label{subsec:limitations}
Real world data is never perfect. The principal limitation of the present, theory-guided article is the fact that it does not yet account for noisy data, which will deteriorate the quality of the learned Hamiltonian. Before SHNNs can be used to extract the behavior of real physical systems (see below), this effect will need to be quantified.

Further, since we are learning a continuous function with a neural network, the dataset has to densely cover the relevant region in the input space (phase space) to obtain a high-quality model. Such dense and vast datasets may not be available in reality. Yet, especially for high-dimensional systems, restricting to small regions where data is available does not inhibit solid results, also when these regions have holes, or even when considering multiple disconnected components.

\subsection{Outlook}
Using symplectic training and corrections of the modified Hamiltonian makes HNNs more powerful, but many further generalizations of our method could be considered. For example, directly learning the time-dependent generating functions $\hatH(p, q, t)$ of Proposition~\ref{prop:2} from a data set $\{(y_0, t_0, y_1, t_1)_i\}$ with variable time steps is an interesting question for future research. Alternatively, generalizations to Poisson systems $\dot y = B(y) \grad H(y)$ (with suitable conditions on the matrix $B(y)$), which model e.g. interactions with electromagnetism or allow to express Hamiltonian mechanics in non-canonical coordinates~\cite[Sec.~VII.2]{GNI}, seem like another fruitful subject.

In conclusion, Symplectic Hamiltonian Neural Networks are a promising ``grey-box'' approach, using physics-priors to build better machine learning algorithms and simultaneously explain why they work. Applications to almost all fields of physics are imaginable, and seem especially exciting in data-rich yet hard-to-model disciplines like the earth's climate or space weather.

%% file: sa-proof.tex

Consider a real smooth Hamiltonian $H : \Omega \rightarrow \RR$ and let $y(t) = (p(t), q(t))$ denote the smooth exact solution of the corresponding Hamilton equation. Hamiltonian Neural Networks as introduced by \citet{greydanus} are asked to predict a smooth scalar function $\hatH : \Omega \rightarrow \RR$.\added{The optimization, however, uses its symplectic in-graph gradient $J^{-1} \grad \hatH$~\eqref{eq:loss-hnn}, numerically calculated with the forward Euler method. As this method is not symplectic, the corresponding ODE, which the neural network tries to learn, is not Hamiltonian~\cite[Thm.~VI.2.6]{GNI}. We can quantify the extent to which the learned ODE is not Hamiltonian, which we expect to realize in the loss during training.}

\added{Concretely, fixing $h > 0$ such that an observation $(y_0, y_1) = (y(0), y(h))$ exists, we ask that the network learn $\hatH$ with the following loss}\footnote{\added{For this analysis,} it is key to consider finite differences on the right-hand side. We do not treat the case where these differences are replaced by true analytic gradients of $H$.}
\begin{equation}\label{ap:eq:hnn-loss}
\begin{split}
	-\grad_q \hatH(p_0, q_0) \stackrel{\mathcal L}{=} \frac{p_1 - p_0}{h} \qq{and} \grad_p \hatH(p_0, q_0) \stackrel{\mathcal L}{=} \frac{q_1 - q_0}{h}.
\end{split}
\end{equation}

\added{After Taylor expanding $y_1$ around $y_0$ and using Hamilton's equations, one obtains the following for the mixed second derivatives of the network:}

\begin{equation}
	\grad_{qp} \hatH - \grad_{pq} \hatH \stackrel{\mathcal L}{=} h \left(\grad_{pq} H \cdot \grad_{qp} H - \grad_{pp} H \cdot \grad_{qq} H\right) \mathrel{\slash\mkern-16.7mu =} 0.
\end{equation}

Noting that any neural network with fully connected linear layers and a smooth activation function (like $\tanh$ in our case) is also smooth, \added{partial derivatives should commute and} this difference should \added{always} be zero. \added{However,} this analysis tells us that standard HNNs will learn a function whose gradients match those of $H$ up to an error (a mismatch) of order $h$. The squared $L^2$ loss \added{of eq.}~\eqref{eq:loss-hnn} will thus be proportional to~$h^2$ \added{when using the forward Euler method}.